\documentclass[runningheads]{llncs}

\usepackage{eccv}

\usepackage{eccvabbrv}

\usepackage{graphicx}
\usepackage[accsupp]{axessibility}  

\usepackage{booktabs}
\usepackage{array}
\usepackage{multirow}
\usepackage{makecell}
\usepackage{tabularx}
\usepackage[table]{xcolor}   
\usepackage{arydshln}        

\usepackage{amsmath,amssymb}

\usepackage{url}

\usepackage{subcaption}

\usepackage{hyperref}
\usepackage{orcidlink}

\begin{document}


\title{PA-VAD: Diffusion-Based \\Pseudo-Only Video Anomaly Detection \\via Domain-Aligned Memory Updates} 

\titlerunning{PA-VAD: Pseudo-Only Video Anomaly Detection}

\author{
Satoshi Hashimoto\thanks{Corresponding author}  \and
Yanan Wang \and
Hitoshi Nishimura \and
Mori Kurokawa
}

\authorrunning{S.~Hashimoto et al.}

\institute{
KDDI Research, Inc., Fujimino, Saitama, Japan\\
\email{\{st-hashimoto,wa-yanan,ht-nishimura,mo-kurokawa\}@kddi.com}
}

\maketitle

\begin{abstract}
Deploying video anomaly detection (VAD) in the real world is often constrained by the scarcity, privacy, and cost of collecting real abnormal footage.
We propose PA-VAD, a novel pseudo-only framework that trains an anomaly detector without using any real abnormal videos, by pairing real normal videos with diffusion-synthesized pseudo-abnormal videos generated from a small set of real normal images. Beyond proposing a generation-driven training pipeline, we make a key empirical discovery: pseudo anomalies exhibit a characteristic spatiotemporal magnitude bias in feature space, which can dominate Multiple Instance Learning and degrade generalization if left unaddressed. To counter this pseudo-induced bias, we introduce the Domain-Aligned Regularized Module (DARM), which combines domain alignment with usage-aware memory updates to balance prototype coverage and stabilize optimization under biased pseudo supervision.
Extensive experiments demonstrate that PA-VAD achieves 98.2\% AUC on ShanghaiTech, 82.5\% on UCF-Crime, and 95.1\% on XD-Violence, and further improves generalization to unseen anomaly classes in open-set evaluations. Notably, PA-VAD surpasses the best real-abnormal WVAD baselines on ShanghaiTech and XD-Violence by +0.6\%  and +0.9\% , respectively, and improves over the UVAD state of the art on UCF-Crime by +1.9\% —showing that high-accuracy VAD is attainable without collecting real abnormal videos.
\end{abstract}

\begin{figure}[!t]
  \centering
  
  \includegraphics[width=\linewidth]{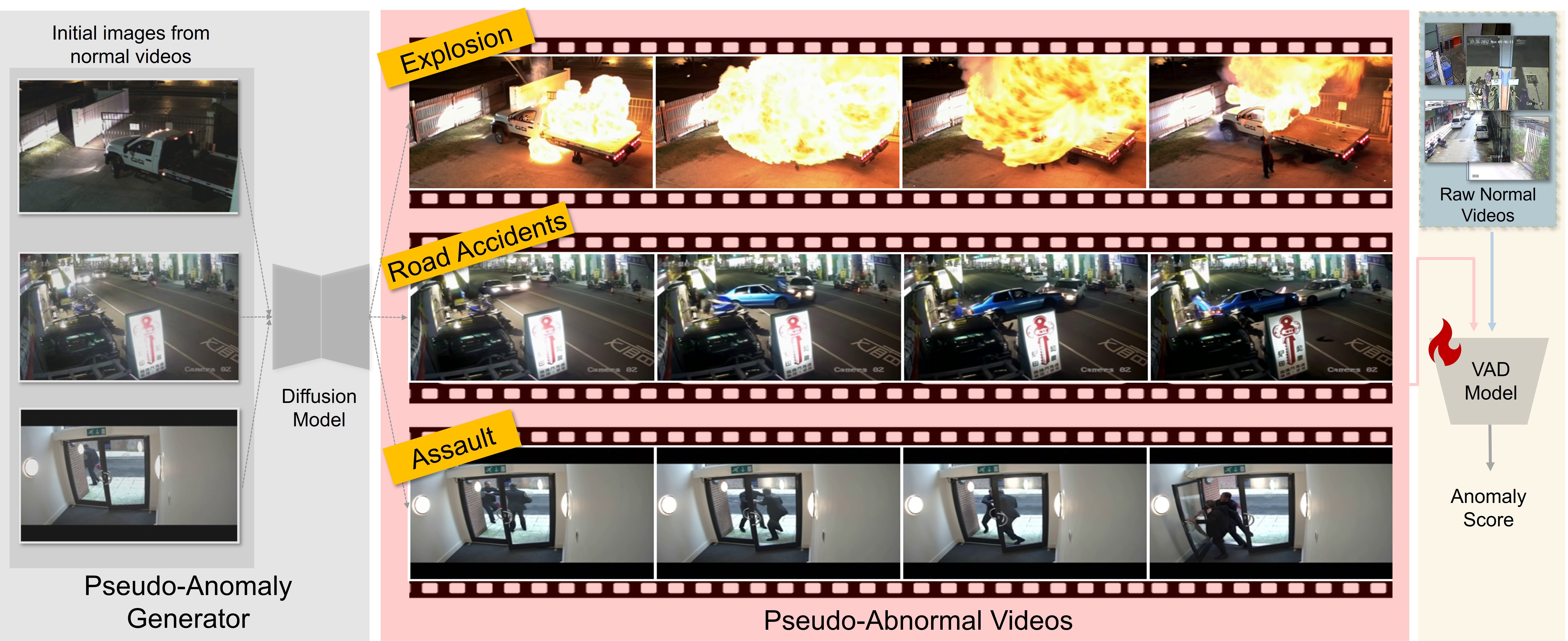}
  \caption{
Our PA-VAD framework generates class-aware pseudo-abnormal videos (e.g., Explosion, Road Accidents, Assault) from a small set of normal images.
These controllable pseudo anomalies provide scalable, diverse supervision, removing the need for real abnormal data and broadening the range of learnable abnormal patterns.}
  \label{fig:1}
\end{figure}

\section{Introduction}
\label{sec:intro}

Ensuring public safety and order has driven extensive research on video anomaly detection (VAD) in computer vision. VAD aims to identify abnormal events such as assaults, traffic accidents, and theft in surveillance videos. Prior works span two main families: (i) unsupervised VAD (UVAD), which trains only on normal data~\cite{ren2015unsupervised,liu2018future,yu2022deep,liu2021hybrid,zaheer2022generative,sun2023lefr,sun2023hsc,zhang2024mgstrl,Delic_2025_ICCV}; (ii) weakly supervised VAD (WVAD), which relies on normal and abnormal videos with video-level labels~\cite{sultani2018real,cho2023cmrl,wu2024vadclip,zhang2023exploiting,URDMU_zh,zhong2019graph,lv2023unbiased,tian2021weakly,Ye2025VERA,Du_2024_CVPR,Zhang2025HolmesVAU,wu2020not}.
Despite rapid progress, deployment remains challenging. UVAD methods can be brittle under distribution shift and label noise, whereas WVAD approaches typically assume access to substantial real abnormal footage, which is costly and often constrained by safety, privacy, or rarity. A complementary direction synthesizes pseudo anomalies to ease data scarcity~\cite{gvvad2025,Rai_2024_CVPRW_VAND}. Yet, in most cases the synthesized clips supplement rather than replace real anomalies, and the synthesis is driven by rule-based edits or narrow generative priors, so the reliance on real abnormal data is reduced but not removed.

\noindent\textbf{Our approach.}
We introduce \textbf{PA-VAD}, a generation-driven framework that learns an anomaly detector from pseudo-anomalous videos synthesized from a small set of real normal images.
Training uses no real abnormal videos, while evaluation follows the standard benchmark splits and protocols.
Unlike prior pseudo-anomaly works that mainly supplement real anomalies and rely on heuristic edits or narrow generators, our synthesis pipeline is designed to replace real anomalies: we select class-relevant seed images with CLIP and perform VLM-based prompt rewriting before driving a video diffusion model, which improves fidelity and scene consistency without manual prompt search.

When training solely on diffusion-synthesized pseudo anomalies without any real abnormal videos, a naïve detector can become biased; we discover that synthesis artifacts (e.g., exaggerated motions, physical implausibility, and temporal inconsistency) often yield abnormally large feature norms---a spatiotemporal magnitude bias---that skews learning toward magnitude cues.
To counter this pseudo-induced bias, we introduce the Domain-Aligned Regularized Module (DARM), combining domain alignment between real Normal and pseudo-Normal streams with usage-aware memory updates to suppress magnitude-driven bias and stabilize learning for stronger anomaly discrimination.

We conduct extensive experiments on multiple benchmarks, showing that our approach outperforms state-of-the-art UVAD baselines and can surpass strong methods that rely on real abnormal videos, while substantially reducing data collection costs and improving open-set generalization to unseen anomalies.

\begin{itemize}
  \item We propose a generation-driven VAD framework that trains without real abnormal videos while evaluating under standard weakly supervised splits, substantially reducing data collection costs.
  \item We propose the Class-Aware Pseudo-Anomaly Generator (CA-PAG), 
  a video diffusion–based pseudo-anomaly generator that attains high-fidelity abnormal videos 
  through CLIP-guided initial image selection and VLM-driven prompt refinement.
  \item We propose the Domain-Aligned Regularized Module (DARM), an adaptive memory module that mitigates the pseudo-induced large-magnitude bias via domain alignment and usage-aware updates, enabling stable MIL training with synthesized anomalies.
  \item Through comprehensive experiments on multiple datasets, we demonstrate state-of-the-art results against UVAD methods and show that our approach can surpass strong methods that depend on real abnormal videos.
\end{itemize}

\begin{figure}[t]
  \centering
  \begin{minipage}[c]{0.5\columnwidth}
    \centering
    \includegraphics[width=\linewidth]{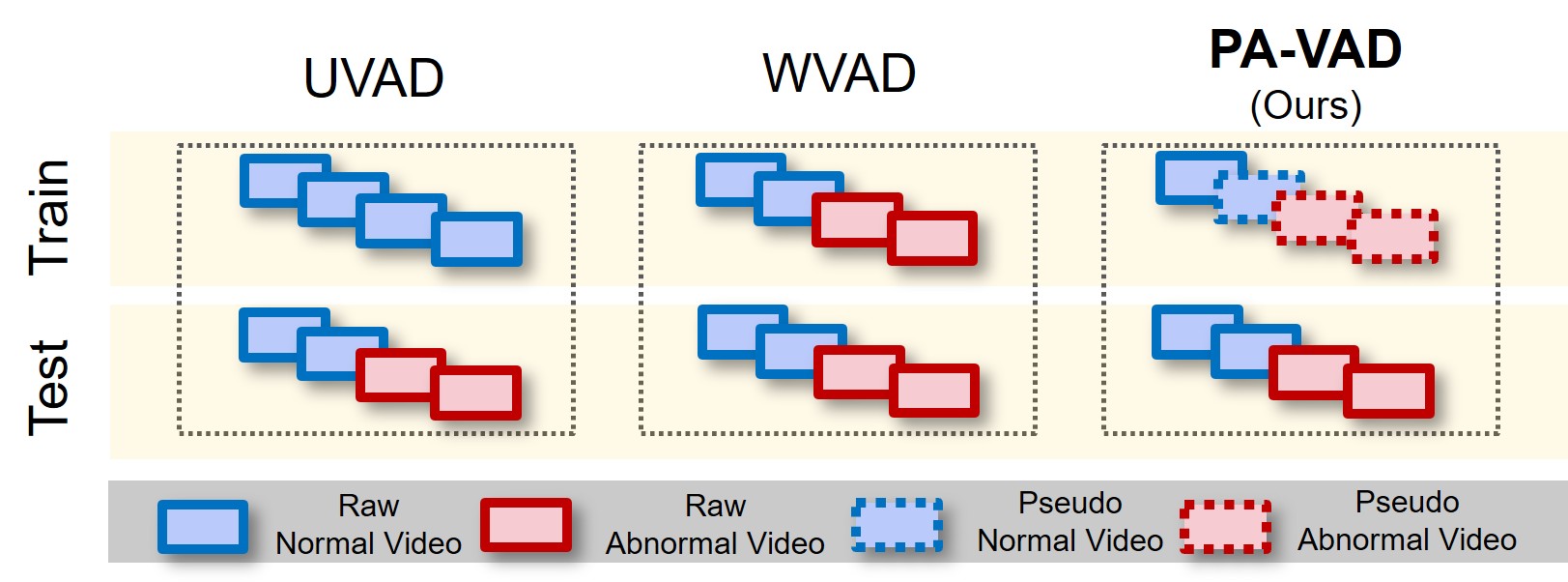}
  \end{minipage}\hfill
  \begin{minipage}[c]{0.45\columnwidth}
    \caption{Overview of our framework. PA-VAD trains on real \textit{Normal} and diffusion-generated pseudo videos (no real \textit{Abnormal}) and is evaluated on standard splits with real \textit{Normal}/\textit{Abnormal}.}
    \label{fig2}
  \end{minipage}
\end{figure}

\begin{figure*}[t]
\centering
\includegraphics[width=\textwidth]{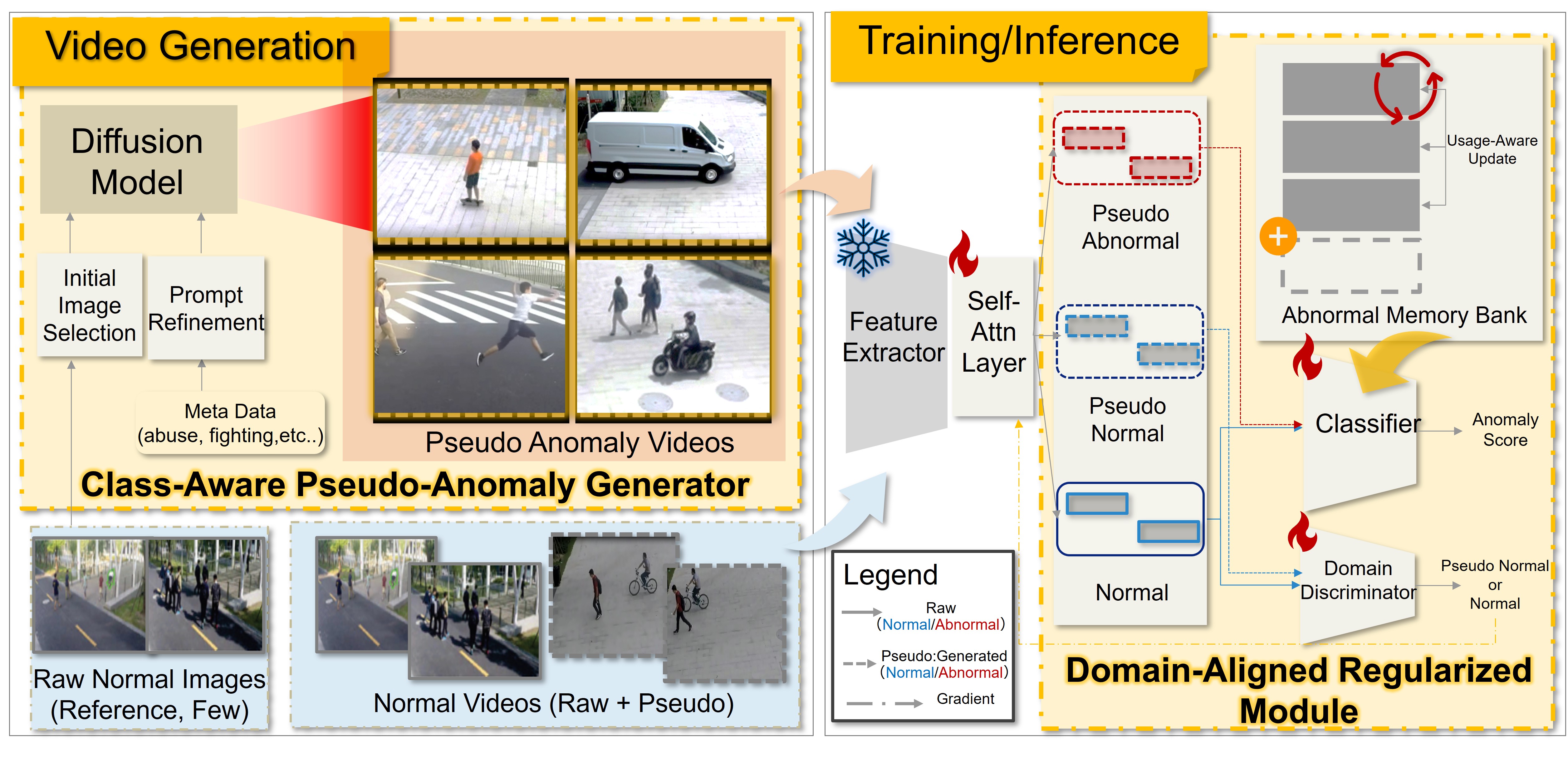}
\caption{Overview of our framework. Starting from a small set of real normal images and class texts, we synthesize pseudo-abnormal videos via an image-to-video diffusion process, then train a classifier on real normal and synthesized pseudo-abnormal videos. A Domain-Aligned Regularized Module—designed to account for the characteristic large spatiotemporal magnitude of synthesized anomalies—mitigates Multiple Instance Learning (MIL) bias and enables accurate detection.}
\label{fig1}
\end{figure*}

\section{Related Work}
\subsection{Video Anomaly Detection}
We group prior work into two families—unsupervised (UVAD) and weakly supervised (WVAD) based methods—and summarize their techniques and limitations.
In UVAD, classical approaches include dictionary learning on normal-only data~\cite{ren2015unsupervised} and prediction/reconstruction paradigms~\cite{liu2018future,yu2022deep,liu2021hybrid,zaheer2022generative,sun2023lefr,sun2023hsc,zhang2024mgstrl}. Recent methods that integrate prediction with reconstruction have proven effective; to explicitly address small-scale anomalies and background interference, MGSTRL learns multi-grained spatio-temporal representations via three proxy tasks—continuity judgment, discontinuity localization, and contrastive missing-frame estimation in feature space—achieving state-of-the-art results, especially on small-scale anomalies~\cite{zhang2024mgstrl}.
A persistent deployment issue for UVAD is the assumption that training covers all normal variations, which rarely holds in the wild. When previously unseen normal patterns arise due to camera/view changes or noise, UVAD methods tend to suffer from false positives.

In contrast, WVAD has become popular as a practical discriminative framework trainable with weak video-level labels.
A large body of work explores architectures using video-level supervision to infer frame-level anomalies~\cite{sultani2018real,cho2023cmrl,wu2024ovvad,wu2024vadclip,zhang2023exploiting,URDMU_zh,zhong2019graph,lv2023unbiased,tian2021weakly,Chen2025SFVAD}.
Sultani \emph{et al.}~\cite{sultani2018real} introduced a MIL-based training scheme under weak labels. Each video is split into segments and represented as a bag of instances: a normal bag $\tilde{B}_{n}=\{\mathbf{f}_n^i\}_{i=1}^{m}$ and an abnormal bag $\tilde{B}_{a}=\{\mathbf{f}_a^i\}_{i=1}^{l}$, where $\mathbf{f}\!\in\!\mathbb{R}^K$ denotes a $K$-dimensional feature vector and $m,l$ are the numbers of segments.
A detector $\mathcal{D}$ outputs an anomaly score $s=\mathcal{D}(\mathbf{f})$ for each segment. Under the MIL assumption, the detector is optimized so that the maximum score in abnormal bags exceeds that in normal bags, using ranking loss:
\begin{equation}
\mathcal{L}_{\text{MIL}}
= \max\!\left(0,\; 1 - \max_{i \in \tilde{B}_a} s_a^i + \max_{i \in \tilde{B}_n} s_n^i \right).
\label{eq:mil_rank}
\end{equation}

Unlike UVAD, such weak supervision learns discriminative representations and tends to be more robust to multi-scene settings and complex backgrounds. WVAD continues to advance, with studies such as Chen \emph{et al.}~\cite{Chen2025SFVAD} extending it by leveraging single-frame supervision. Recently, WVAD methods have adopted vision–language models (VLMs) for better interpretability. Du \emph{et al.}~\cite{Du_2024_CVPR} target causal understanding of abnormal events using hard/soft prompts. Zhang \emph{et al.}~\cite{Zhang2025HolmesVAU} fine-tune VLM on large captioned datasets and report leading performance. However, these WVAD approaches typically presuppose large-scale collection of real abnormal videos, which can be costly and constrained by safety and privacy, limiting practical deployment.

\subsection{Pseudo Video Generation}
Several works have explored generating pseudo data to boost VAD.
Cai \emph{et al.}~\cite{gvvad2025} integrate video generation into weakly supervised training by synthesizing text-conditioned pseudo anomalies mixed with real abnormal videos. This alleviates but does not eliminate reliance on real anomalies, as full real sets remain used during training.
Rai \emph{et al.}~\cite{Rai_2024_CVPRW_VAND} generate pseudo anomalies for UVAD by inpainting frames with a pretrained diffusion model and perturbing optical flow, but their design still limits the diversity and realism of synthesized anomalies.

Recent advances in diffusion models have produced powerful video generators. Two major conditions are common: text-to-video (T2V), which directly synthesizes videos from text prompts, and image-to-video (I2V), which extends a static image over time while preserving spatial structure under a prompt.
Wan~2.2 is a diffusion-Transformer framework with noise-stage mixture of experts and highly compressed latent representations~\cite{wan2025}.
Stable Video Diffusion establishes a three-stage curriculum spanning text–image pretraining, video pretraining, and high-quality video finetuning~\cite{blattmann2023svd}.
While these models are strong general-purpose generators, they are not tailored for anomaly synthesis; prompting to obtain rare abnormal behaviors is challenging and typically requires careful search and conditioning.

\section{Proposed Method}
\label{sec:method}
We propose \textbf{PA-VAD}, a pseudo-only framework that trains a video anomaly detector \emph{without any real abnormal videos}. 
As illustrated in Fig.~\ref{fig2}, training uses real \textit{Normal} videos together with diffusion-generated pseudo-abnormal videos, while evaluation follows the standard splits with real \textit{Normal}/\textit{Abnormal}.
An overview is provided in Fig.~\ref{fig1}.
PA-VAD comprises two components: (i) \emph{Class-Aware Pseudo-Anomaly Generator} and (ii) \emph{Domain-Aligned Regularized Module (DARM)}. We detail each module below.

\subsection{Class-Aware Pseudo-Anomaly Generator}
The goal is to synthesize pseudo-abnormal videos from only a small set of real normal images and class metadata (abnormal class names) by driving a video diffusion model. We adopt an image-to-video (I2V) inference process conditioned on (a) an initial image and (b) a textual prompt. A central challenge is the covariate shift between synthesized anomalies and real data: prior synthesis for VAD either mixes generated clips with real abnormal videos~\cite{gvvad2025} or perturbs real images/videos~\cite{Rai_2024_CVPRW_VAND}, which keeps the domain gap relatively small. In contrast, because our aim is to eliminate reliance on real abnormal footage altogether, \emph{reducing this gap while preserving quality} is crucial. We therefore introduce the Class-Aware Pseudo-Anomaly Generator (CA-PAG), 
whose synthesis is controlled by two levers: (1) \textbf{initial image selection} to anchor the generation near the target class domain, and (2) \textbf{prompt refinement} to tailor textual conditioning to the chosen initial image via zero-shot vision–language reasoning. 

\begin{figure}[t]
  \centering
  \begin{subfigure}[t]{0.49\columnwidth}
    \centering
    \includegraphics[width=\linewidth]{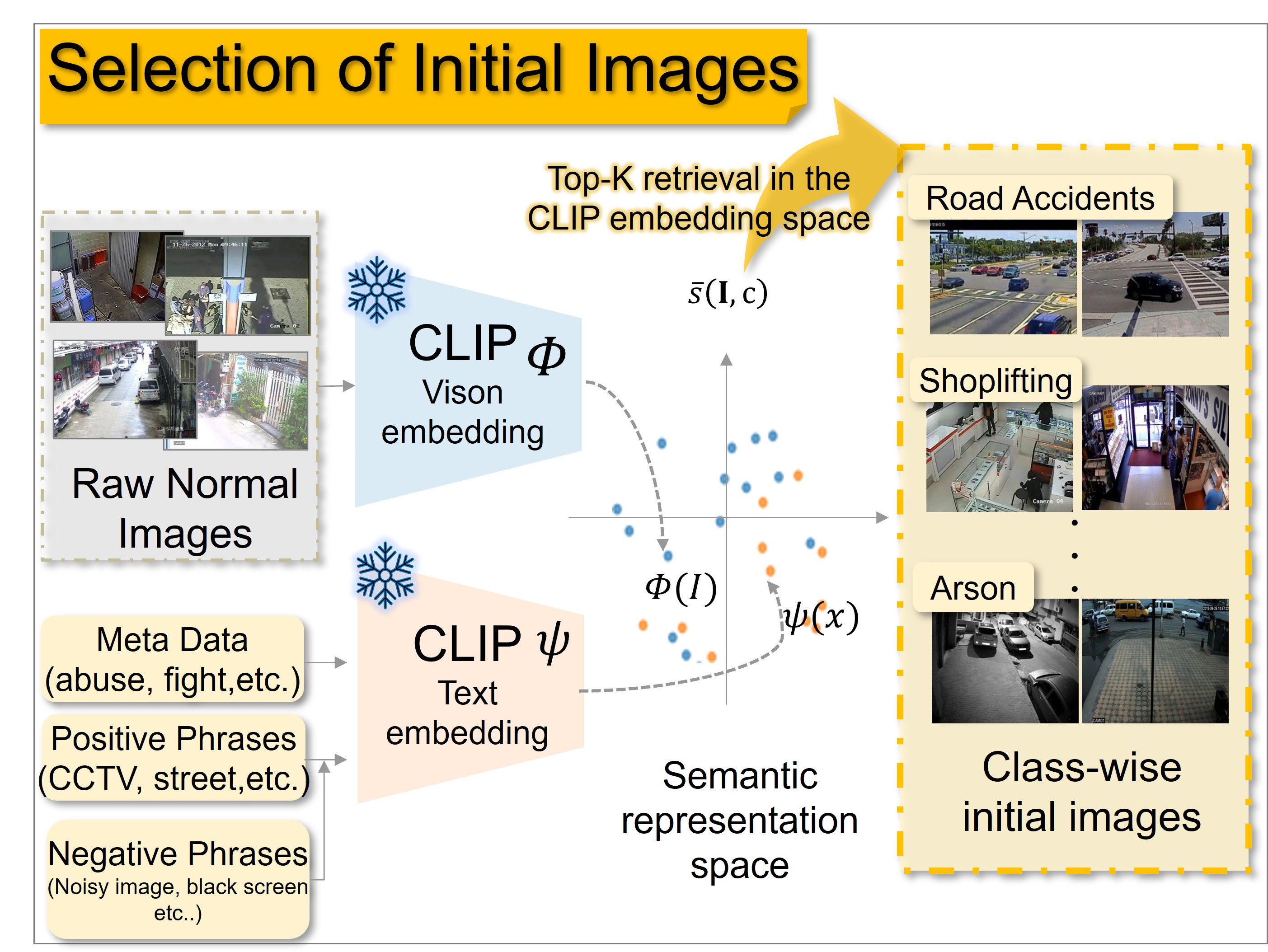}
    \caption{Initial image selection in the vision--text space. We score normal images using a class text with positive phrases and subtract a negative similarity, taking the Top-$K$ per class.}
    \label{fig:4a}
  \end{subfigure}
  \hfill
  \begin{subfigure}[t]{0.49\columnwidth}
    \centering
    \includegraphics[width=\linewidth]{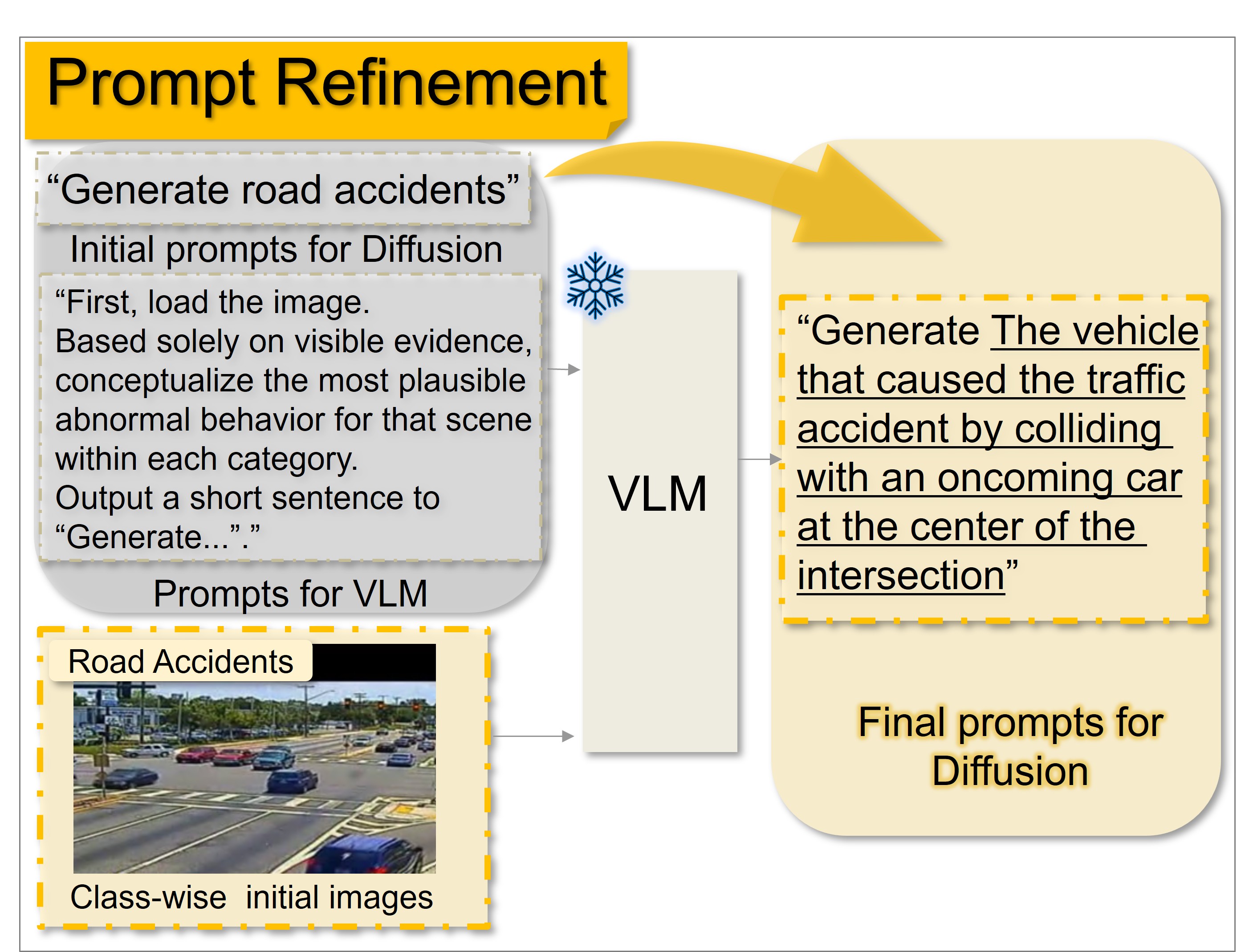}
    \caption{Prompt refinement. A VLM converts initial-image cues into concise, class-consistent abnormal descriptions, concatenating a template before driving the diffusion model.}
    \label{fig:4b}
  \end{subfigure}
  \caption{Overview of our class-aware pseudo-anomaly generation pipeline.}
  \label{fig:4}
\end{figure}

\noindent\textbf{Initial Image Selection.}
Naively sampling initial images at random from the normal set risks large domain gaps for certain classes (e.g., selecting an indoor home scene for the target class ``Road Accident'' in UCF-Crime~\cite{sultani2018real}). Instead, we propose to select \emph{class-relevant} inits with CLIP~\cite{radford2021clip} using a Top-$K$ strategy in the joint vision–text space (Fig.~\ref{fig:4} (a)).

Concretely, given a class label $x_c$, we augment it with positive phrases $g_{\mathrm{pos}}$ to enforce surveillance-style attributes such as camera footage and CCTV, and introduce a set of negative phrases $g_{\mathrm{neg}}$ to suppress nuisances such as black screens or channel logos. Let $\Phi$ and $\psi$ denote the CLIP image and text encoders. Please refer to the supplementary material for the detailed description of each phase. Define unit-normalized embeddings
$\hat{\mathbf v}(\mathbf I)=\Phi(\mathbf I)/|\Phi(\mathbf I)|_2$ and $\hat{\mathbf t}(u)=\psi(u)/|\psi(u)|_2$.
We form a single positive text by concatenating the class name with positives,$t_c=\mathrm{concat}(x_c,\, g_{\mathrm{pos}})$
and compute the positive similarity:
\begin{equation}
s_{\mathrm{pos}}(\mathbf I,c)=\big\langle \hat{\mathbf v}(\mathbf I),\, \hat{\mathbf t}(t_c)\big\rangle .
\label{eq:clip_pos}
\end{equation}
We compute each negative similarity:
\begin{equation}
s_{\mathrm{neg}}(\mathbf I)
=
\big\langle
\hat{\mathbf v}(\mathbf I),\,
\hat{\mathbf t}(g_{\mathrm{neg}})
\big\rangle .
\label{eq:clip_neg}
\end{equation}
The final score balances class affinity and nuisance suppression,
\begin{equation}
\tilde{s}(\mathbf I,c)= s_{\mathrm{pos}}(\mathbf I,c)-\lambda\, s_{\mathrm{neg}}(\mathbf I),\qquad \lambda\in[0,1],
\label{eq:clip_final}
\end{equation}
and select the initial set $\mathrm{TopK}_c$ for class $c$ using a scene-balanced ranking scheme.
We apply scene balancing to prevent seed selection from being dominated by populous camera/location scenes. Specifically, with $s$ denoting a camera/location scene, we subsample the normal-image pool with per-scene budgets proportional to $\mathrm{count}(s)^{\alpha}$ and prioritize the highest-scoring candidate from each scene before filling the remaining Top-$K$ slots by the global ranking $\tilde{s}(\mathbf I,c)$.

\noindent\textbf{Prompt Refinement for Pseudo Anomalies}
Using a raw class name as a prompt is simple but often ambiguous, which harms generation consistency and text–video alignment~\cite{mehrabi2023tab,hu2023tifa}. Exhaustive manual search, however, is impractical. We therefore refine prompts with a vision–language model (VLM) by extracting objects and scene cues from each initial image and turning them into concise, class-consistent abnormal descriptions (Fig.~\ref{fig:4} (b)).

Given initial images $\{\mathbf{I}_c^{(k)}\}_{k=1}^{K}$ for class $c$, a class text $x_c$, and a refinement instruction $g_{\mathrm{vlm}}$, the VLM produces short candidate phrases:
\begin{equation}
\hat{s}_c^{(k)} \;=\; \mathcal{G}_{\varphi}\!\bigl(\mathbf{I}_c^{(k)},\, g_{\mathrm{vlm}},\, x_c\bigr).
\end{equation}
To standardize camera and rendering attributes and suppress artifacts, each phrase is concatenated with a set of template prompts $g_{\mathrm{tem}}$ (e.g., \textit{natural movement, fixed camera}):
\begin{equation}
p_c^{(k)} \;=\; \mathrm{concat}\!\bigl(\hat{s}_c^{(k)},\, g_{\mathrm{tem}}\bigr).
\end{equation}
Finally, the video diffusion model $\mathcal{G}_{\theta}$ is conditioned on the refined prompt and the corresponding initial image to synthesize a pseudo-abnormal clip:
\begin{equation}
\tilde{\mathbf{V}}_c^{(k)} \;=\; \mathcal{G}_{\theta}\!\bigl(p_c^{(k)},\, \mathbf{I}_c^{(k)}\bigr).
\end{equation}
Please see the supplementary material for details of the refinement instruction.

\subsection{Domain-Aligned Regularized Module}
This module trains and infers an anomaly detector using synthesized pseudo-abnormal videos $\tilde{\mathbf{V}}$ and real normal videos $\mathbf{V}$. When we simply plug synthesized videos into existing weakly supervised pipelines, we observe a characteristic failure mode: generated clips tend to exhibit excessive motion and viewpoint changes, inflating feature norms and biasing MIL Top-$k$ optimization toward a few high-magnitude pseudo instances. This bias induces a covariate shift at test time against real anomalies and degrades sensitivity. Table~\ref{tab:magnitude_comparison} reports the mean I3D feature norms on UCF-Crime for real vs.\ pseudo anomalies.

To counter this and exploit the diversity of synthesized data, we introduce the Domain-Aligned Regularized Module (DARM) with two components:   
(\textup{i}) \textbf{Domain alignment:} a DANN-based objective~\cite{ganin2016dann} that aligns real and pseudo normal distributions to reduce covariate shift; 
(\textup{ii}) \textbf{Usage-aware memory update:} an update rule that pulls under-utilized slots toward their responsibility-weighted centers to rebalance coverage across abnormal modes.
The domain alignment term is applied to mean-pooled features of real normal and pseudo-normal streams. 
By encouraging domain-invariant representations via gradient reversal, it reduces magnitude shifts between real and pseudo data, preventing pseudo anomalies from dominating MIL Top-$k$ due to inflated norms.
The usage-aware memory update balances the abnormal prototypes. 
It computes per-slot usage from soft assignments and updates under-utilized slots toward their responsibility-weighted centers, mitigating the tendency of a few high-norm slots to monopolize MIL Top-$k$ selection.
We follow the UR-DMU\cite{URDMU_zh} classifier, which provides a GL-MHSA–based encoder, dual (normal/abnormal) memory banks for prototype separation, and a MIL scoring head with uncertainty control.
Adding the DARM objectives on top of this backbone effectively suppresses the MIL bias caused by excessive spatiotemporal magnitude in synthesized anomalies while preserving representation diversity across abnormal prototypes.

\begin{table}[t]
\centering
\caption{Mean $\ell_2$ norms of frozen backbone features extracted from real and pseudo anomalies on UCF-Crime. Feature spaces differ across backbones.}
\label{tab:magnitude_comparison}
\scriptsize
\setlength{\tabcolsep}{6pt}
\renewcommand{\arraystretch}{1.05}
\begin{tabular}{lccc}
\toprule
 & I3D\cite{carreira2017quo} & Qwen\cite{bai2025qwen2_5_vl} & C3D\cite{tran2015learning} \\
\midrule
Real   & 20.52 & 24.22 & 19.14 \\
Pseudo & 23.03 & 25.55 & 20.41 \\
\bottomrule
\end{tabular}
\end{table}

\smallskip\noindent\textbf{Formulation.}
Given a video, a feature extractor $\Phi_{\mathrm{vid}}$ produces segment features $\mathbf{f}=\Phi_{\mathrm{vid}}(\mathbf{V})\in\mathbb{R}^{T\times D}$. We embed them with temporal convolution and self-attention to obtain $\tilde{\mathbf{f}}\in\mathbb{R}^{B\times T\times d}$:
\begin{equation}
\tilde{\mathbf{f}}=\mathrm{SelfAttn}\!\bigl(\mathrm{Temporal}(\mathbf{f})\bigr).
\end{equation}

Let $\mathbf{M}_A,\mathbf{M}_N\in\mathbb{R}^{K\times d}$ denote abnormal and normal memory banks. Their memory guided augmentations are $\mathbf{h}_A(\tilde{\mathbf{f}})$ and $\mathbf{h}_N(\tilde{\mathbf{f}})$. We concatenate these with $\tilde{\mathbf{f}}$ and predict framewise anomaly scores $\hat{\mathbf{a}}$:
\begin{equation}
\hat{\mathbf{a}}=\sigma\!\Bigl(\mathbf{W}\,\bigl[\,\tilde{\mathbf{f}}\,;\,\mathbf{h}_A(\tilde{\mathbf{f}})+\mathbf{h}_N(\tilde{\mathbf{f}})\,\bigr]+\mathbf{b}\Bigr).
\end{equation}

Let $\mathcal{S}_{\mathrm{abn}}$ be indices of abnormal embeddings and 
$\tilde{\mathbf{f}}_{\mathrm{abn}}=\tilde{\mathbf{f}}[\mathcal{S}_{\mathrm{abn}}]$. 
Unfolding over time yields $\mathbf{Z}\in\mathbb{R}^{(B_a\,T)\times d}$. 
Define row-wise $\ell_2$ normalization by 
$\mathbf{D}_Z=\operatorname{diag}(\|\mathbf{Z}_1\|_2,\ldots,\|\mathbf{Z}_{B_aT}\|_2)$ 
and $\bar{\mathbf{Z}}=\mathbf{D}_Z^{-1}\mathbf{Z}$. 
Similarly, define
\begin{equation}
\mathbf{D}_M=\operatorname{diag}\!\bigl(\|\mathbf{m}_1\|_2,\ldots,\|\mathbf{m}_K\|_2\bigr),
\end{equation}
and let $\bar{\mathbf{M}}_A=\mathbf{D}_M^{-1}\mathbf{M}_A$.

With temperature $\tau>0$, we define the assignment matrix and slot usage as:

\begin{equation}
\mathbf{Q} = \operatorname{softmax}\!\left(\frac{\bar{\mathbf{Z}}\bar{\mathbf{M}}_A^{\!\top}}{\tau}\right)
   \in \mathbb{R}^{(B_a\,T)\times K}.
\end{equation}
\begin{equation}
\mathbf{u} = \frac{1}{B_a\,T}\,\mathbf{1}^{\top}\mathbf{Q}
   \in \mathbb{R}^{K}.
\end{equation}

\emph{(i) Domain alignment}
We align real Normal and pseudo-Normal distributions through adversarial training.
Let $\bar{\mathbf f}_N$ and $\bar{\mathbf f}_{\tilde N}$ denote the mean features of Normal and pseudo-Normal streams,
and $D(\cdot)$ the domain discriminator with gradient reversal layer $G_{\lambda_{\mathrm{da}}}(\cdot)$.
The domain alignment loss is
\begin{equation}
\begin{aligned}
\mathcal{L}_{\mathrm{DA}}
&= \mathrm{BCE}\big(D(G_{\lambda_{\mathrm{da}}}(\bar{\mathbf f}_N)),\, y_N\big) \\
&\quad + \mathrm{BCE}\big(D(G_{\lambda_{\mathrm{da}}}(\bar{\mathbf f}_{\tilde N})),\, y_{\tilde N}\big) \\
&\quad + \lambda_{\mathrm{dist}} \big\|\bar{\mathbf f}_N - \bar{\mathbf f}_{\tilde N}\big\|_2^2,
\end{aligned}
\end{equation}
where $\lambda_{\mathrm{da}}\!\ge\!0$ controls the gradient reversal strength
and $\lambda_{\mathrm{dist}}\!\ge\!0$ weights the explicit Normal–pseudo-Normal feature alignment term.

\emph{(ii) Usage-aware update.}
While domain alignment adjusts global statistics, 
some slots may remain underutilized. 
Let $\boldsymbol{\mu}_k=\big(\sum_n Q_{nk}\mathbf{Z}_n\big)\big/\big(\sum_n Q_{nk}\big)$ 
be the responsibility-weighted center of slot $k$, 
$\mathbf{m}_k$ the $k$-th row of $\mathbf{M}_A$, 
and $\bar{u}=\tfrac{1}{K}\sum_k u_k$ the mean slot usage. 
We encourage balanced coverage by pulling underused slots more strongly toward their centers:
\begin{equation}
\mathcal{L}_{\mathrm{upd}}
=\tfrac{1}{K}\sum_{k=1}^{K}
\Bigl(\tfrac{\bar{u}}{u_k+\varepsilon}\Bigr)^{\!\beta}
\!\bigl\|\,\mathbf{m}_k-\boldsymbol{\mu}_k\,\bigr\|_2^2.
\end{equation}

\smallskip\noindent\textbf{Objective.}
Our final loss augments the UR\mbox{-}DMU discrimination term with the two regularizers:
\begin{equation}
\mathcal{L}=\mathcal{L}_{\mathrm{UR\mbox{-}DMU}}+\lambda_{1}\mathcal{L}_{\mathrm{DA}}+\lambda_{2}\mathcal{L}_{\mathrm{upd}}.
\end{equation}

\section{Experiments}
\subsection{Experimental Setup}
\noindent
\textbf{Setup.}
Unless otherwise specified, we follow an \emph{unsupervised training} regime: the training data include only real normal videos and the names of abnormal classes that might appear at inference time; \emph{no real abnormal videos} are used for training. At evaluation, both real normal and real abnormal videos are included, and we report frame-level metrics.

\noindent
\textbf{Datasets.}
We evaluate on three widely used benchmarks with distinct characteristics: ShanghaiTech (SHT)~\cite{liu2018future}, UCF-Crime (Crime)~\cite{sultani2018real} and XD-Violence (XD)~\cite{wu2020not}.
SHT consists of fixed-view campus surveillance scenes with diverse illumination and viewpoints, spanning 13 locations and 437 video clips, and covering 14 abnormal categories (e.g., bicycle riding, skateboard intrusion, fighting), with 63 abnormal clips available in the training split. Following common practice for weakly supervised evaluation\cite{tian2021weakly,cho2023cmrl,lv2023unbiased}, we use the revised split of~\cite{zhang2019temporal} at test time, while keeping training \emph{abnormal-free}.
For SHT, we synthesize $10$ pseudo-abnormal clips per class (total $140$) and additionally generate $50$ pseudo-normal clips for domain alignment.
Crime is a large-scale collection of real-world surveillance videos (about 1{,}900 videos, $\sim$128 hours) covering 13 abnormal classes (e.g., explosion, abuse), with 810 abnormal clips in the training split.
For Crime, we synthesize $20$ pseudo clips per class (total $260$) and additionally generate $80$ pseudo-normal clips. Following prior setups\cite{URDMU_zh}, we apply 10-crop augmentationl.
XD is a large-scale multi-scene violence benchmark with 4{,}754 untrimmed videos and 217 hours, spanning 6 violence categories.
We follow the official weakly supervised split with 3{,}954 training videos and 800 testing videos.
For XD, we synthesize 10 pseudo-abnormal clips for each of 6 classes and 50 pseudo-normal clips for domain alignment; details of pseudo-generation are provided in the supplementary material. Pseudo-video synthesis is performed once offline; generating 60 clips takes $\sim$10 hours on two RTX 6000 Ada GPUs (96GB).

\noindent
\textbf{Metrics.}
To assess the \emph{generation quality} of the synthesized videos, we report standard video generation metrics. For image-level quality, we compute Fr\'echet Inception Distance (FID) and Kernel Inception Distance (KID) between frame distributions of generated and real videos. For spatiotemporal fidelity, we report Fr\'echet Video Distance (FVD) and Kernel Video Distance (KVD) based on I3D features, capturing motion and temporal coherence. For \emph{anomaly detection}, we follow the VAD literature and report frame-level Area Under the ROC Curve (AUC). In all quantitative comparisons, scores of prior methods are directly quoted from their original papers unless otherwise stated.

\noindent
\textbf{Implementation Details.}
We use the Wan~2.2 image-to-video diffusion model (I2V-A14B)~\cite{wan2025} with a resolution of $832\times480$ and a frame length of $81$. For prompt refinement, we employ the Qwen3 30B-A3B vision–language model~\cite{qwen3technicalreport}. As an auxiliary video feature extractor (for analysis/metrics), we use the final projector embedding (3{,}584-D) of Qwen2.5-VL 7B-Instruct~\cite{bai2025qwen2_5_vl}. CLIP ViT-B/32~\cite{radford2021clip} is used for initial image selection. Our implementation is based on PyTorch and HuggingFace Transformers.

\begin{figure}[!t]
  \centering
  \begin{minipage}[c]{0.62\columnwidth}
    \centering
    \includegraphics[width=\linewidth,height=0.26\textheight,keepaspectratio]{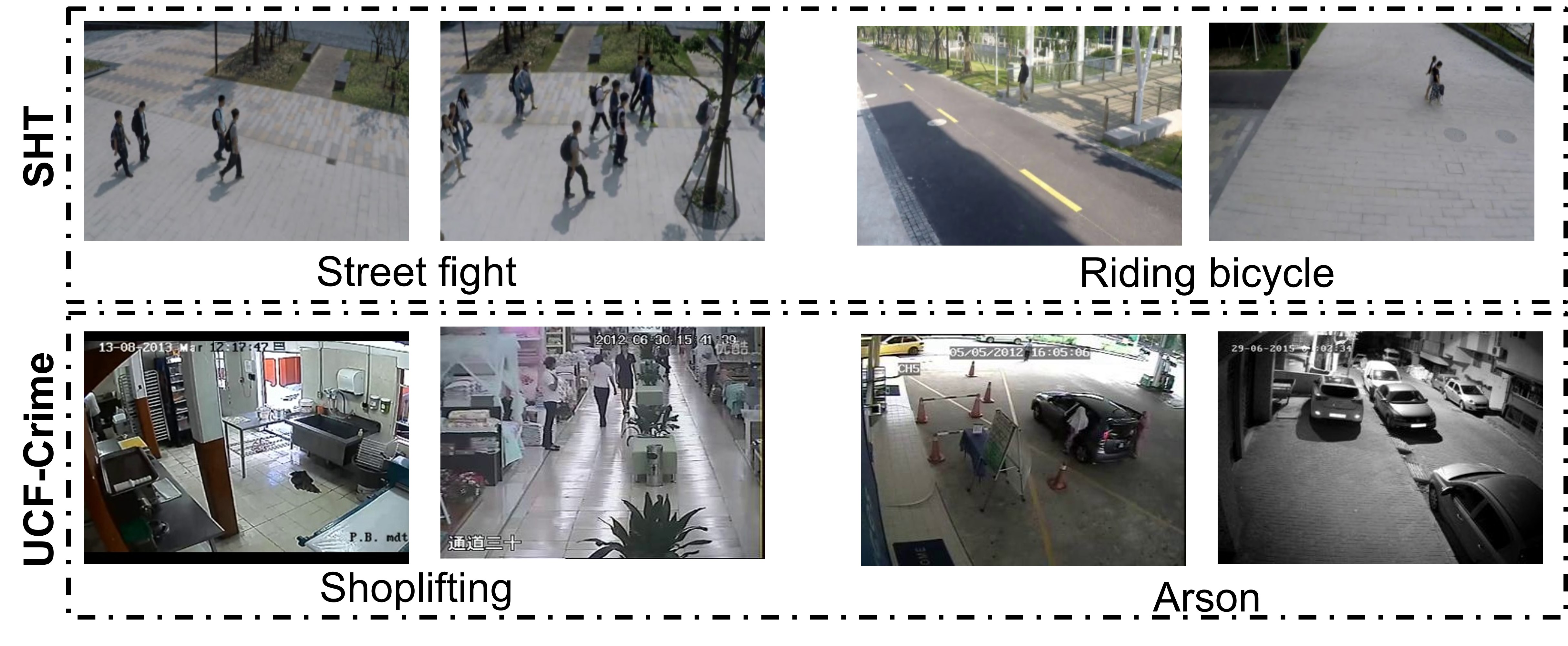}
  \end{minipage}\hfill
  \begin{minipage}[c]{0.35\columnwidth}
    \caption{Initial image examples produced by our CLIP-based selection for SHT and UCF-Crime. Images are class-relevant and scene-consistent, serving as anchors for I2V synthesis.}
    \label{fig:seed_results}
  \end{minipage}
\end{figure}

\noindent
\textbf{Initial Images and Prompt Design.}
Using the init selection process in Sec.~3.1, we obtain class-relevant init images for SHT and Crime (Fig.~\ref{fig:seed_results}). They exhibit minimal semantic mismatch to target classes. Representative examples of VLM-based prompt refinement are shown in Table~\ref{tab:prompt_refinement}, where the refined text better reflects the scene context of the initial image. These inits and refined prompts are then fed to the I2V generator to construct the pseudo-anomaly datasets. Please refer to the supplementary material for the detail.

\begin{table}[!t]
\centering
\caption{Examples of prompt refinement by the VLM. }
\label{tab:prompt_refinement}
\scriptsize
\setlength{\tabcolsep}{2pt}
\renewcommand{\arraystretch}{1.05}
\begin{tabular}{l >{\arraybackslash}m{9cm}}
\toprule
w/o refine & Generate Shoplifting \\
\rowcolor{gray!12}
w/ refine  & Generate Store theft involves an individual removing merchandise from shelves, concealing it in a backpack, and evading surveillance cameras. \\
\bottomrule
\end{tabular}
\end{table}

\begin{figure*}[!t]
\centering
\includegraphics[width=\textwidth]{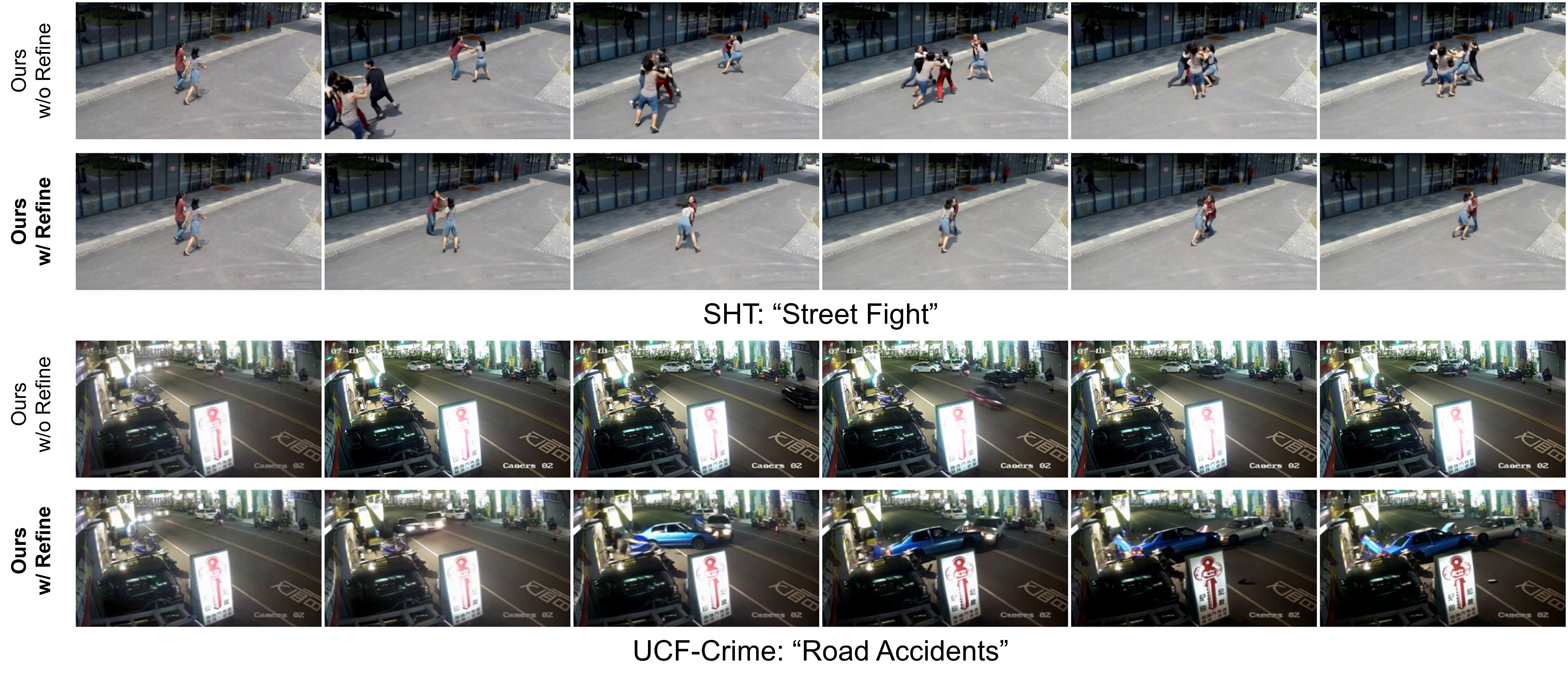}
\caption{Qualitative examples of synthesized pseudo anomalies. }
\label{fig:6}
\end{figure*}

\begin{figure*}[!h]
\centering
\includegraphics[width=\textwidth]{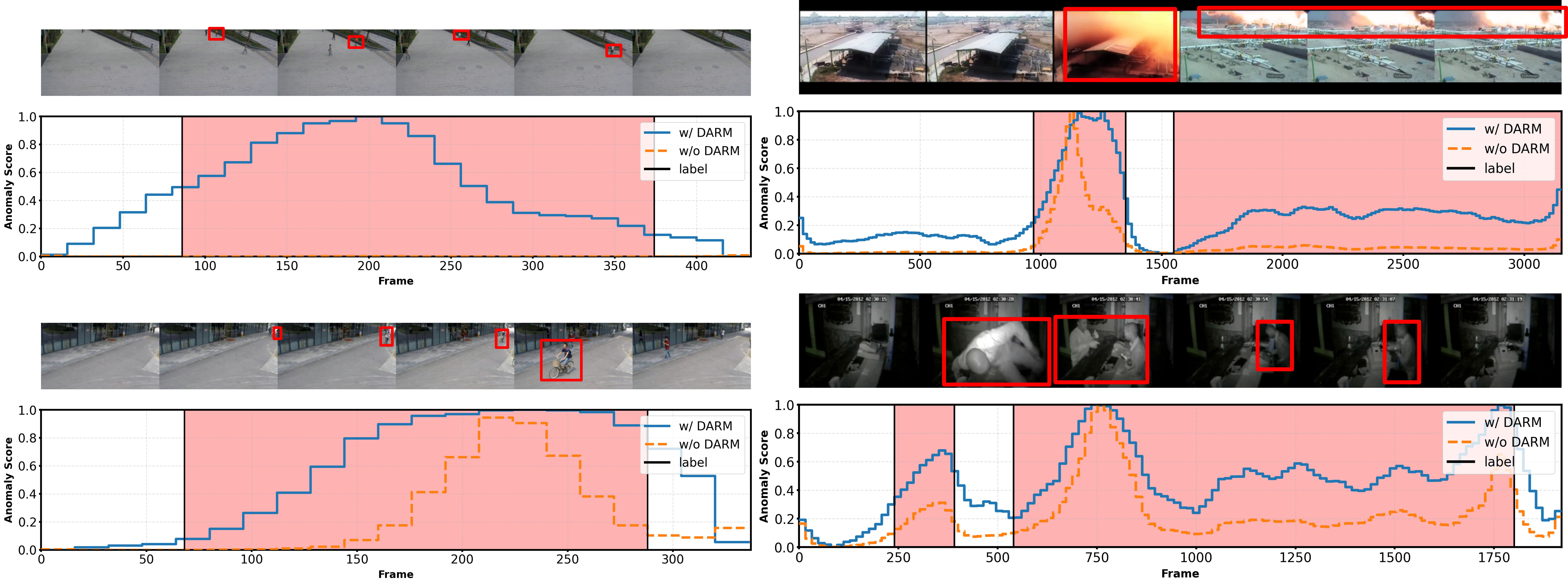}
\caption{Qualitative detection results. Left: SHT (top:skateboarder, bottom:biker). Right: UCF-Crime (top:explosion,bottom:burglary). The adaptive memory module mitigates MIL bias and improves sensitivity to subtle abnormal segments.}
\label{fig:7}
\end{figure*}

\subsection{Qualitative Evaluation}
\textbf{Class-Aware Pseudo-Anomaly Generator (CA-PAG)}
Fig.~\ref{fig:6} shows examples of generated videos.
Without prompt refinement (\textit{w/o Refine}, i.e., using the raw class label), ambiguous instructions degrade consistency and text–video alignment, leading in SHT to artifacts such as the intrusion of large fighting crowds, object merging/disappearance across frames, and excessive motion that ignores physical plausibility. On Crime, we likewise observe multiple vehicles appearing/disappearing.
With prompt refinement (\textit{w/ Refine}), the VLM-guided prompts produce scenes that reflect the initial image context: on SHT we observe person fighting present in the initial image with more natural motion; on Crime we observe an inter-vehicle collision without disappearance/merging artifacts.

\noindent
\textbf{Domain-Aligned Regularized Module(DARM).}
Fig.~\ref{fig:7} shows score traces (vertical: score; horizontal: time).
Left: in ``skateboard intrusion,'' \textit{w/ DARM} responds sharply at abnormal segments, while \textit{w/o DARM} barely rises.
In the ``biker'' case, \textit{w/ DARM} suppresses false positives and concentrates its peak within the annotated interval.
Right: in ``explosion,'' \textit{w/ DARM} sustains higher sensitivity during the burning phase.
In ``burglary,'' it produces temporally tighter peaks aligned with abnormal actions. These examples demonstrate that DARM not only enhances responsiveness to true abnormal events but also suppresses magnitude-driven false negatives, yielding temporally sharper and more semantically aligned detection peaks.

\subsection{Quantitative Evaluation}
\textbf{Pseudo-anomaly videos quality.}
Table~\ref{tab:pseudo} reports SHT generation quality.
As prompt refinement is applied, all metrics improve. In particular, we observe a reduction of FVD by 97 points and KVD by 22.7 points, indicating better motion and temporal coherence. These gains are consistent with the qualitative trends in Fig.~\ref{fig:6}, suggesting that refined prompts enhance physical plausibility and scene consistency. In contrast, KID remains unchanged. As it reflects per-frame appearance rather than temporal dynamics, both settings share identical image statistics, while CA-PAG improves motion coherence captured by FVD/KVD.

\noindent\textbf{DARM for detection.}
Table~\ref{tab:4} reports frame-level AUC against prior UVAD and WVAD methods under the Real/Pseudo setting (i.e., training without any real abnormal videos).
On SHT, PA-VAD with DARM achieves $98.2\%$ AUC, surpassing the best UVAD method~\cite{zhang2024mgstrl} and even outperforming the strongest WVAD competitor CMRL~\cite{cho2023cmrl} ($97.6\%$) by \textbf{+0.6} points.
On XD-Violence, it attains $95.1\%$ AUC and exceeds the strongest Real/Real baseline UR-DMU~\cite{URDMU_zh} ($94.2\%$) by \textbf{+0.9} points.
These results indicate that DARM mitigates pseudo-induced bias and improves transfer across diverse anomaly domains.
On Crime, our Real/Pseudo setting is naturally disadvantaged against Real/Real
baselines, yet PA-VAD with DARM remains competitive, and still surpasses the
UVAD state-of-the-art~\cite{zhang2024mgstrl} ($80.6\%$) by \textbf{+1.9} points.
UVAD methods are the closest comparison in abnormal-data access, though the
level of supervision is not strictly equal, as PA-VAD additionally relies on
large pretrained generative and vision-language priors; this trades
hard-to-collect real abnormal footage for readily available pretrained
knowledge---a practical advantage for real-world deployment.
Within our PA-VAD framework, replacing the plain UR-DMU~\cite{URDMU_zh} with
DARM consistently improves performance under the same setting, from
$96.0\%\!\to\!98.2\%$ on SHT, $80.2\%\!\to\!82.5\%$ on Crime, and
$92.5\%\!\to\!95.1\%$ on XD; a consistent gain also holds for the Sultani
et al. \cite{sultani2018real} classifier, indicating that domain alignment and the usage-aware update are
important to fully exploiting pseudo anomalies without overfitting to their
magnitude bias.

\noindent\textbf{Ablation of CA-PAG and DARM.} 
Table~\ref{tab:ablation_capag_darm} disentangles the individual contributions of the CA-PAG and DARM components on SHT, allowing us to quantify the effect of each design choice.
Starting from a random-init baseline ($86.7\%$), CLIP-based initial image selection raises AUC to $94.9\%$, a substantial gain of $+8.2$ points, confirming the benefit of anchoring generation to class-relevant normal seeds.
Enabling prompt refinement on top of init image selection already improves AUC from $94.9\%$ to $96.0\%$, indicating that class-aware textual conditioning is important even without any regularization on the detector side.
Within DARM, the usage-aware memory update is the primary driver of gains: adding the update term on top of CA-PAG boosts AUC to $97.6$–$97.7\%$, whereas domain alignment alone provides only modest improvement.
The best result ($98.2\%$) is obtained when domain alignment and usage-aware update are combined, supporting our design that DARM suppresses the magnitude bias by reducing the real/pseudo discrepancy and preventing high-magnitude prototypes from dominating MIL optimization.

\begin{table}[!t]
\centering
\caption{Quality of pseudo-anomaly videos.}
\label{tab:pseudo}
\scriptsize
\setlength{\tabcolsep}{4pt}
\renewcommand{\arraystretch}{1.05}
\begin{tabular}{lcccc}
\toprule
Data & FVD$\downarrow$ & FID$\downarrow$ & KVD$\downarrow$ & KID$\downarrow$ \\
\midrule
Ours w/o Refine  & 701  & 83.9 & 57.4  & \textbf{0.03} \\
\rowcolor{gray!12}
Ours w/ Refine   & \textbf{604}  & \textbf{78.0} & \textbf{34.7} & \textbf{0.03} \\
\bottomrule
\end{tabular}
\end{table}

\begin{table*}[!t]
\centering
\caption{
Frame-level AUC (\%) comparison on ShanghaiTech (SHT), UCF-Crime (Crime), and XD-Violence (XD). ``Feature'' denotes the frozen clip-level backbone used by the detector; ``--'' indicates methods without such features. ``Data type (Nor/Abn)'' denotes the source of normal/abnormal training videos (real or pseudo).}

\label{tab:4}
\footnotesize
\setlength{\tabcolsep}{5pt}
\renewcommand{\arraystretch}{1.12}
\resizebox{\textwidth}{!}{%
\begin{tabular}{ c l l c c c c c }
\toprule
Protocol & Methods & Reference & Feature &
\makecell{Data type\\(Nor/Abn)} &
SHT & Crime  & XD\\
\midrule
\multirow{4}{*}{UVAD}
& LERF~\cite{sun2023lefr}                      & AAAI23 & --  & Real/-     & 78.6 & -- & --   \\
& HSC~\cite{sun2023hsc}                        & CVPR23 & --  & Real/-     & 83.4 & --& --  \\
& MGSTRL~\cite{zhang2024mgstrl}                & CVPR24 & --  & Real/-     & 87.5 & 80.6& --  \\
& SeeKer~\cite{Delic_2025_ICCV}                & ICCV25 & --  & Real/-     & 85.5 & --  & --  \\
\hdashline
\multirow{4}{*}{WVAD}
& Sultani et al.~\cite{sultani2018real}        & CVPR18 & C3D          & Real/Real  & -- & 75.4 & -- \\
& CMRL~\cite{cho2023cmrl}                      & CVPR23 & I3D & Real/Real  & 97.6 & 86.1 & --  \\
& UR-DMU~\cite{URDMU_zh}                       & AAAI23 & I3D          & Real/Real  & --   & 87.0 & 94.2 \\
& VERA~\cite{Ye2025VERA}                       & CVPR25 & InternVL2 & Real/Real  & --   & 86.6 & 88.3 \\
\hdashline
\multirow{3}{*}{\makecell{PA\mbox{-}VAD\\(Ours)}}
& {Ours w/Sultani~\cite{sultani2018real}}
& {}
& {Qwen2.5-VL}
& {Real/Pseudo}
& {93.7}
& {77.8}
& {81.8}
\\
& {Ours w/UR-DMU~\cite{URDMU_zh}}
& {}
& {Qwen2.5-VL}
& {Real/Pseudo}
& {96.0}
& {80.2}
& {92.5}
\\
& \cellcolor{gray!12}{Ours w/DARM}
& \cellcolor{gray!12}{}
& \cellcolor{gray!12}{Qwen2.5-VL}
& \cellcolor{gray!12}{Real/Pseudo}
& \cellcolor{gray!12}{\textbf{98.2}}
& \cellcolor{gray!12}{\textbf{82.5}}
& \cellcolor{gray!12}{\textbf{95.1}}
\\
\bottomrule
\end{tabular}
}
\end{table*}

\begin{table*}[!t]
\centering
\caption{Ablation studies of PA-VAD.}
\label{tab:ablation_combined}
\begin{subtable}[t]{0.48\textwidth}
\centering
\caption{Effect of CA-PAG and DARM components on SHT.}
\label{tab:ablation_capag_darm}
\scriptsize
\setlength{\tabcolsep}{4pt}
\renewcommand{\arraystretch}{1.10}
\begin{tabular}{ccccc}
\toprule
\multicolumn{2}{c}{CA-PAG} & \multicolumn{2}{c}{DARM} & \multirow{2}{*}{AUC(\%)} \\
\cmidrule(lr){1-2}\cmidrule(lr){3-4}
Initial & Refine & DA & Update & \\
\midrule
 &  &  &  & 86.7 \\
\checkmark &  &  &  & 94.9 \\
\checkmark & \checkmark &  &  & 96.0 \\
\checkmark &  &  & \checkmark & 97.6 \\
\checkmark & \checkmark &  & \checkmark & 97.7 \\
\checkmark &  & \checkmark &  & 96.0 \\
\checkmark & \checkmark & \checkmark &  & 96.8 \\
\rowcolor{gray!12}
\checkmark & \checkmark & \checkmark & \checkmark & \textbf{98.2} \\
\bottomrule
\end{tabular}
\end{subtable}
\hfill
\begin{subtable}[t]{0.48\textwidth}
\centering
\caption{Mean $\ell_2$ norm of real/pseudo abnormal features in the detector space.}
\label{tab:darm_norm}
\scriptsize
\setlength{\tabcolsep}{5pt}
\renewcommand{\arraystretch}{1.6}
\resizebox{\textwidth}{!}{%
\begin{tabular}{l c c c}
\toprule
Feature norm & SHT & Crime & XD \\
\midrule
Real Abn.              & 9.13   & 11.06  & 39.08  \\
\hdashline
Pseudo Abn. w/o DARM   & 199.48 & 52.94  & 420.10 \\
Pseudo Abn. w/ DA      & 183.50 & 45.87  & 262.40 \\
\rowcolor{gray!12}
Pseudo Abn. w/ DARM    & \textbf{8.39} & \textbf{12.80} & \textbf{33.63} \\
\bottomrule
\end{tabular}%
}
\end{subtable}
\end{table*}

\begin{table}[!t]
\centering
\caption{Effect of the number of synthesized clips on SHT. We vary the number of generated pseudo-abnormal clips (total = 140) and report the frame-level AUC (\%).}
\label{tab:num_clips}
\scriptsize
\setlength{\tabcolsep}{4pt}
\renewcommand{\arraystretch}{1.05}
\begin{tabular}{lcccccc}
\toprule
Clips & 14 & 35 & 70 & 105 & 140 & 175 \\
\midrule
SHT (\%) & 85.4 & 94.6 & 96.9 & 97.3 & \textbf{98.2} & 97.7 \\
\bottomrule
\end{tabular}
\end{table}

\noindent\textbf{Magnitude bias analysis.} To verify that DARM mitigates bias on the \emph{anomaly} side, we measure the
mean $\ell_2$ norm of real and pseudo abnormal features in the detector space
(Table~\ref{tab:darm_norm}).
The mild gap in the frozen backbone (cf. Table~\ref{tab:magnitude_comparison})
is strongly amplified where MIL Top-$k$ operates: without DARM, the
pseudo-abnormal norm inflates to over $20\times$ the real-abnormal norm on SHT
($199.48$ vs.\ $9.13$), forming a magnitude shortcut.
Comparing the last two rows of Table~\ref{tab:darm_norm} isolates the
usage-aware update: domain alignment alone (w/ DA) only partially narrows the
gap, as it aligns the normal streams rather than the abnormal prototypes,
whereas full DARM restores the pseudo-abnormal norms to the real-abnormal range
across all datasets. This confirms the usage-aware update as the key component for correcting the bias on the abnormal side.

\noindent\textbf{Effect of the number of pseudo clips.}  Table~\ref{tab:num_clips} further analyzes robustness to the amount of synthesized data on SHT.
Increasing the number of pseudo-abnormal clips from $14$ to $140$ steadily improves performance from $85.4\%$ to $98.2\%$ AUC, and we already reach $96.9\%$ with only $70$ clips.
Even when the number of synthesized clips is reduced to around the scale of the $63$ real abnormal clips in the standard WVAD split, PA-VAD with DARM remains highly competitive.
When we further increase the number of synthesized clips to $175$, the AUC slightly drops to $97.7\%$, suggesting a mild saturation effect where overly many pseudo clips introduce redundancy or noise that does not further benefit, and can slightly harm, generalization.
Overall, the results indicate that pseudo-only training can match or exceed Real/Real pipelines, offering a practical option when abnormal data is unavailable.

\noindent\textbf{Further Analysis with Related Methods.} We compare PA-VAD with pseudo-data generation methods and the single-frame-supervised SF-VAD/FPL. Table~\ref{tab:related_quant} reports two comparisons: prior pseudo-data results from~\cite{gvvad2025} and the reported SF-VAD/FPL results. Cai \emph{et al.}~\cite{gvvad2025} mix generated pseudo anomalies with real abnormal videos; in contrast, PA-VAD uses no real abnormal videos. PA-VAD remains competitive with SF-VAD/FPL on SHT ($98.2$ vs.\ $98.4$) despite not relying on the single precise abnormal-frame annotations that SF-VAD uses. Note that SF-VAD differs from PA-VAD in its baseline, as it employs single frame-level supervision together with real abnormal videos.

\begin{table}[!t]
\centering
\caption{Comparison with pseudo-data generation methods and SF-VAD.}
\label{tab:related_quant}
\scriptsize
\setlength{\tabcolsep}{3pt}
\renewcommand{\arraystretch}{1.13}

\scalebox{0.90}{%
\begin{tabular}{c l c c c c c}
\toprule
\multirow{2}{*}{Protocol} & \multirow{2}{*}{Method}
& \multirow{2}{*}{Feature}
& \multicolumn{2}{c}{Data type}
& \multirow{2}{*}{SHT} & \multirow{2}{*}{Crime} \\
\cmidrule(lr){4-5}
& & & Nor. & Abn. & & \\
\midrule

\multirow[c]{2}{*}{WVAD}
& Cai et al.~\cite{gvvad2025} w/RTFM
& \multirow[c]{2}{*}{CLIP}
& \multirow[c]{2}{*}{Real+Pseudo}
& \multirow[c]{2}{*}{Real+Pseudo}
& \multirow[c]{2}{*}{--}
& 87.3 \\
& Cai et al.~\cite{gvvad2025} w/LAP
& & & & & 89.3 \\
\hdashline

SF-VAD
& SF-VAD/FPL~\cite{Chen2025SFVAD}
& I3D RGB
& Real & Real
& 98.4 & 89.9 \\
\hdashline

\rowcolor{gray!12}
PA-VAD
& PA-VAD (ours)
& Qwen2.5-VL
& Real & Pseudo
& \textbf{98.2} & \textbf{82.5} \\
\bottomrule
\end{tabular}%
}
\end{table}

\noindent\textbf{Generalization to Unseen Anomalies.}
\label{sec:open_set}
A key concern in pseudo-only training is overfitting to anomaly categories used for pseudo generation, potentially harming robustness to unseen anomalies.
Following OpenVAD's open-set protocol~\cite{zhu2022towards} on Crime and XD, we generate pseudo-abnormal videos from a restricted subset of anomaly classes (\emph{seen}) and evaluate on the full test set including held-out classes (\emph{unseen}).
We repeat three random class splits and report mean AUC with standard deviation for PA-VAD; for prior methods, we report the values from their original papers.

Tables~\ref{tab:open_ucf} and~\ref{tab:open_xd} show that PA-VAD consistently outperforms OpenVAD across all settings, even with only one seen class (XD: $72.50 \to 89.08$, \textbf{+$16.6$} points), indicating transfer beyond seen-class artifacts.
We attribute this open-set robustness primarily to DARM.
As discussed in the paper, DARM suppresses pseudo-induced shortcut cues (e.g., feature-magnitude bias) that can dominate MIL-based training and lead to brittle decision rules tied to synthesized artifacts.
More importantly for open-set transfer, DARM's \emph{usage-aware memory update} acts as an anti-overfitting mechanism against seen-class specialization:
it reweights memory updates by slot usage so that slots repeatedly activated by the same seen-class--specific modes are updated less, while under-used slots are promoted.
This prevents memory from being monopolized by a few seen-driven prototypes and encourages the detector to rely on more transferable, mode-agnostic abnormal evidence, consistent with the improved sensitivity to unseen classes under open-set splits.

\begin{table*}[t]
\centering
\begin{minipage}[t]{0.56\textwidth}
\centering
\caption{Open-set on Crime.}
\label{tab:open_ucf}
\setlength{\tabcolsep}{3.6pt}
\renewcommand{\arraystretch}{1.08}
\resizebox{\linewidth}{!}{%
\begin{tabular}{lcccc}
\toprule
\textbf{Seen anom.} & 1 & 3 & 6 & 9 \\
\midrule
RTFM~\cite{tian2021weakly}   & 75.91 & 76.98 & 77.68 & 79.55 \\
OpenVAD~\cite{zhu2022towards} & 76.73 & 77.78 & 78.82 & 80.14 \\
\rowcolor{gray!15}
Ours &
$\mathbf{77.65}_{\pm\,\mathbf{3.29}}$ &
$\mathbf{80.86}_{\pm\,\mathbf{2.03}}$ &
$\mathbf{82.18}_{\pm\,\mathbf{0.25}}$ &
$\mathbf{82.30}_{\pm\,\mathbf{0.14}}$ \\
\bottomrule
\end{tabular}}
\end{minipage}
\hfill
\begin{minipage}[t]{0.40\textwidth}
\centering
\caption{Open-set on XD.}
\label{tab:open_xd}
\setlength{\tabcolsep}{4.2pt}
\renewcommand{\arraystretch}{1.08}
\resizebox{\linewidth}{!}{%
\begin{tabular}{lcc}
\toprule
\textbf{Seen anom.} & 1 & 4 \\
\midrule
OpenVAD~\cite{zhu2022towards} & 72.50 & 88.25 \\
\rowcolor{gray!15}
Ours &
$\mathbf{89.08}_{\pm\,\mathbf{4.52}}$ &
$\mathbf{94.99}_{\pm\,\mathbf{0.13}}$ \\
\bottomrule
\end{tabular}}
\end{minipage}
\end{table*}

\noindent\textbf{Limitation.}
On UCF-Crime, our method does not surpass the SoTA that trains with real abnormal videos under the WVAD setup.
A likely factor is the difficulty of synthesizing minute-long, context-heavy anomalies (e.g., extended temporal dependencies), which remain challenging for current video generators; such anomalies are more prevalent in UCF-Crime than in SHT, making the dataset inherently harder under a pseudo-only setting.

\section{Conclusion}
We presented PA-VAD, a pseudo-only video anomaly detection framework with two components: CA-PAG for synthesizing class-consistent pseudo anomalies and DARM for mitigating pseudo-induced magnitude bias and stabilizing MIL-based learning.  
Without using any real abnormal footage, PA-VAD achieves competitive or superior performance to strong baselines on SHT, Crime, and XD, demonstrating that high-quality pseudo anomalies with regularized training provide a practical alternative to Real/Real pipelines.  
Future work includes improving motion priors for pseudo generation and extending evaluation to broader video generative models and settings.

\bibliographystyle{splncs04}
\bibliography{main}

\titlerunning{PA-VAD: Pseudo-Only Video Anomaly Detection}
\authorrunning{S.~Hashimoto et al.}

\title{Supplementary Materials for\\PA-VAD: Diffusion-Based Pseudo-Only Video Anomaly Detection via Domain-Aligned Memory Updates}

\author{Satoshi Hashimoto\thanks{Corresponding author} \and
Yanan Wang \and
Hitoshi Nishimura \and
Mori Kurokawa}

\institute{KDDI Research, Inc., Fujimino, Saitama, Japan\\
\email{\{st-hashimoto,wa-yanan,ht-nishimura,mo-kurokawa\}@kddi.com}}

\maketitle

\section*{S1. Additional Details of CA-PAG}

\subsection*{S1.1 Initial image retrieval}

For each abnormal class $c$, we retrieve an initial image from the training set of real normal videos that is visually compatible with the target abnormal context.
Concretely, we first construct a positive query sentence $q_c^{+}$ that describes a class-relevant surveillance scene, and a negative query $q_c^{-}$ that filters out non-surveillance artefacts such as noisy images or logos.
Example phrases of "road accidents" include:
\begin{itemize}
  \item Positive: ``road accidents, roadway, multiple vehicles, street lanes, intersection, traffic lights, road shoulder, median strip, daytime or dusk, fixed high pole camera, moderate motion blur''
  \item Negative: ``dashboard ui overlay, speedometer close-up, racing circuit''
\end{itemize}

We encode all candidate normal frames and both queries using CLIP, and compute similarity between each frame and $q_c^{+}$ while penalizing similarity to $q_c^{-}$.
Figure~1 in the supplementary material visualizes examples of retrieved init images for "road accidents" class of UCF-Crime\cite{sultani2018real}.
This CLIP-based retrieval anchors the synthesis process to class-consistent, surveillance-style scenes without requiring any real abnormal footage.

\begin{figure}[!htbp]
  \centering
  \includegraphics[width=\columnwidth]{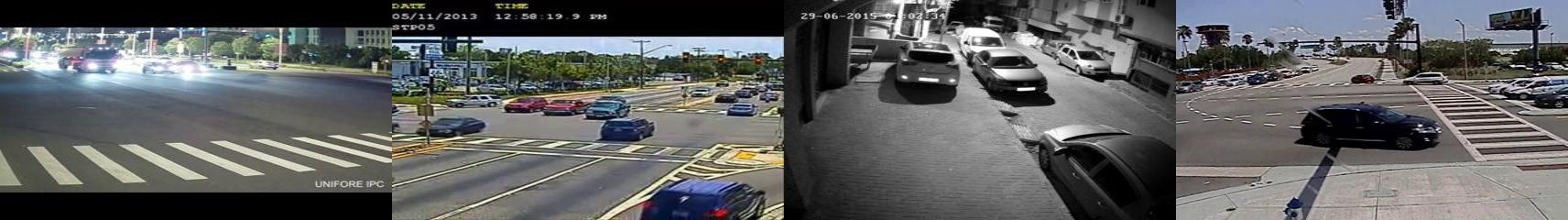}
  \caption{Top-4 retrieved init images for the class ``road accident''. 
  Ranking is left to right (1st$\rightarrow$4th).}
  \label{fig:top4_init}
\end{figure}

\subsection*{S1.2 Prompt refinement with a VLM}

Starting from a coarse abnormal class name (e.g., \textit{``fighting''}, \textit{``robbery''}), we construct an initial English prompt that describes the abnormal event in a surveillance setting.
Given an init image $\mathbf{x}$ and a coarse textual description $p^{\text{coarse}}_c$, we query the VLM with an instruction that (i) asks for a more detailed, temporally grounded description of the abnormal event, (ii) explicitly constrains the camera to be static, and (iii) enforces scene consistency with the given surveillance image. A simple example is as follows:

\begin{quote}
You are an expert in visual scene understanding and anomaly-behavior design.
Given an input image and an anomaly category, describe the most plausible abnormal behavior consistent with both the scene and the category.
Your output must be a single short sentence completing the pattern
``Generate abuse behavior, manifested as \dots''.
Use only elements clearly visible in the image.
Do not invent unseen objects, persons, weapons, or motions.
Keep the action realistic, scene-grounded, and visually justified.
Camera motion is strictly prohibited.
\end{quote}

We provide the full VLM instruction prompt (written in Chinese) used for textual refinement in the supplemental material. The VLM outputs a refined prompt $p^{\text{refined}}_c(\mathbf{x})$ that is used to condition the video diffusion model.
Table~1 shows examples of coarse vs.\ refined prompts.
Typically, the refined prompt adds temporal structure (e.g., gradual approach, interaction, and departure), clarifies the role of normal bystanders, and explicitly states the absence of camera motion, which stabilizes the synthesized videos.

\begin{table}[!htbp]
\centering
\caption{Examples of prompt refinement used in CA-PAG. Original prompts were written in Chinese before translation and refinement.}
\scriptsize
\renewcommand{\arraystretch}{1.20}
\setlength{\tabcolsep}{6pt}
\begin{tabular}{p{3.2cm} p{3.8cm}}
\toprule
\textbf{Before Refinement} & \textbf{After Refinement} \\
\midrule

Generate \textbf{abuse}.&
Generate a person in an office space aggressively shoves and restrains another individual at close distance. \\

\midrule

Generate \textbf{burglary}. &
Generate a person reaches behind a store counter to rummage through items while another stands watching from the front. \\

\midrule

Generate \textbf{robbery}. &
Generate in a parking lot, an individual approaches near a vehicle and forcibly grabs an item from another person. \\

\bottomrule
\end{tabular}
\end{table}

\subsection*{S1.3 Pseudo-anomaly generation with Wan2.2}

We use an image-to-video diffusion model (Wan2.2) to synthesize pseudo-abnormal clips conditioned on the init image and the refined prompt.
Unless otherwise stated, we use a resolution of \texttt{832}$\times$\texttt{480} pixels, \texttt{81} frames per clip, a frame rate of \texttt{16} fps, and \texttt{25} sampling steps.These settings are shared across classes for both ShanghaiTech and UCF-Crime.
For the classifier-free guidance scale, we use \texttt{(3.5,3.5)} for SHT and \texttt{(6.5,4.5)} for UCF-Crime and XD-Violence.

Figure~2 illustrates representative synthesized clips for four classes (burglary, explosion, fighting, moving car, vandalism) and one failure case (shoplifting).
We observe that short, visually localized anomalies (e.g., explosions around a single object, brief collisions, or localized vehicle motion) are synthesized with high fidelity and maintain a consistent surveillance viewpoint.
In contrast, long-horizon, context-heavy crimes such as shoplifting or subtle abuse are more difficult: the generator often produces largely normal behavior with only vague or missing abnormal cues.
We explicitly treat such failure cases as part of the pseudo-anomaly distribution; DARM is designed to tolerate these imperfections and to prevent a few high-magnitude but unfaithful clips from dominating MIL optimization.

\begin{figure*}[t]
    \centering

    \begin{subfigure}[b]{0.95\textwidth}
        \centering
        \includegraphics[width=\linewidth]{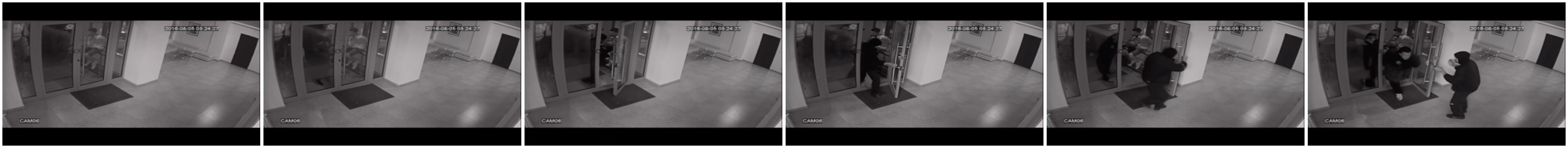}
        \caption{Burglary (UCF-Crime)}
    \end{subfigure}
    \vspace{10pt}

    \begin{subfigure}[b]{0.95\textwidth}
        \centering
        \includegraphics[width=\linewidth]{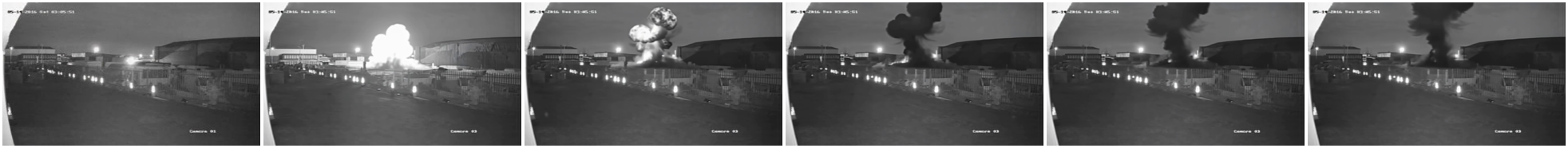}
        \caption{Explosion (UCF-Crime)}
    \end{subfigure}
    \vspace{10pt}

    \begin{subfigure}[b]{0.95\textwidth}
        \centering
        \includegraphics[width=\linewidth]{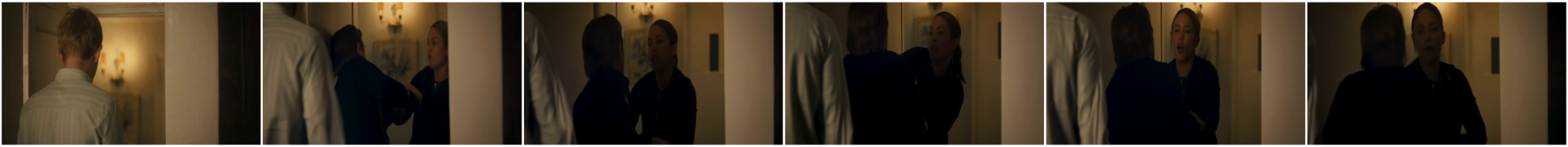}
        \caption{Fighting (XD-Violence)}
    \end{subfigure}
    \vspace{10pt}

    \begin{subfigure}[b]{0.95\textwidth}
        \centering
        \includegraphics[width=\linewidth]{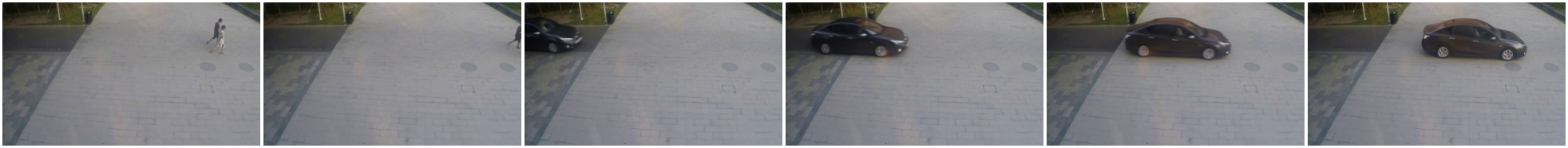}
        \caption{Moving car (ShanghaiTech)}
    \end{subfigure}
    \vspace{10pt}

    \begin{subfigure}[b]{0.95\textwidth}
        \centering
        \includegraphics[width=\linewidth]{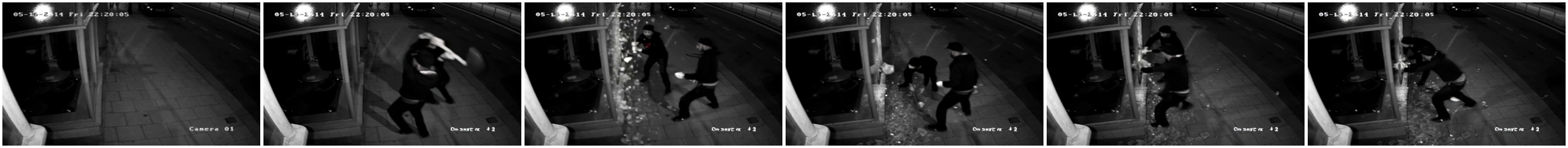}
        \caption{Vandalism (UCF-Crime)}
    \end{subfigure}
    \vspace{10pt}

    \begin{subfigure}[b]{0.95\textwidth}
        \centering
        \includegraphics[width=\linewidth]{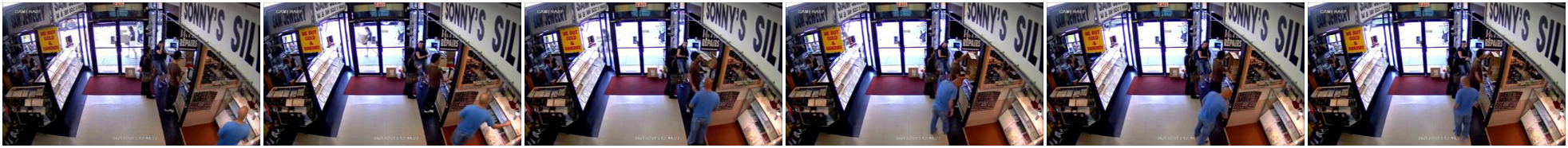}
        \caption{Shoplifting (UCF-Crime, failure example)}
    \end{subfigure}

    \caption{
    Representative synthesized pseudo-anomaly examples used in PA-VAD. 
    Four high-fidelity cases (burglary, explosion, fighting, moving car, vandalism) and one failure case (shoplifting), where the abnormal cue remains vague or partially missing.
    }
    \label{fig:syn_examples_vertical}
\end{figure*}

\section*{S2. Additional Details of DARM}

\subsection*{S2.1 Loss Terms and Coefficients}

Our final training loss follows Eq.~(15) of the main paper:
\[
\mathcal{L}
= \mathcal{L}_{\text{UR-DMU}}
+ \lambda_{1}\mathcal{L}_{\text{DA}}
+ \lambda_{2}\mathcal{L}_{\text{upd}} ,
\]
where each coefficient corresponds to:

\begin{itemize}
  \item $\lambda_{1}$: weight for the domain-alignment loss $\mathcal{L}_{\text{DA}}$ (Eq.~(13))
  \item $\lambda_{2}$: weight for the usage-aware memory update loss $\mathcal{L}_{\text{upd}}$ (Eq.~(14))
  \item $\lambda_{\text{da}}$: gradient reversal strength in Eq.~(13)
  \item $\lambda_{\text{dist}}$: explicit Normal--pseudo-Normal alignment term in Eq.~(13)
  \item $\beta$: exponent controlling the scaling of usage-aware correction in Eq.~(14)
\end{itemize}

All hyperparameters used in the main experiments are summarized in Table~\ref{tab:train_hparams}.

\begin{table*}[!htbp]
\centering
\caption{Training hyperparameters for SHT, UCF-Crime and XD-Violence. All experiments use the same backbone, data pipeline, and MIL configuration; only dataset-dependent constants differ.}
\label{tab:train_hparams}
\renewcommand{\arraystretch}{1.15}
\begin{tabular}{l c c c}
\toprule
\textbf{Hyperparameter} & \textbf{ShanghaiTech} & \textbf{UCF-Crime} & \textbf{XD-Violence}\\
\midrule
Batch size & 4 & 32& 4 \\
Learning rate & $1\times10^{-4}$ & $1\times10^{-5}$ & $1\times10^{-5}$\\
Segment number & $5$ & $64$ & $4$\\
Optimizer & Adam & Adam & Adam\\
weight decay& $1\times10^{-5}$ & $1\times10^{-5}$ & $1\times10^{-5}$\\
\midrule
$\lambda_{1}$ (Eq.~13, weight of $\mathcal{L}_{\mathrm{DA}}$) & 1.0 & 1.0& 1.0 \\
$\lambda_{2}$ (Eq.~14, weight of $\mathcal{L}_{\mathrm{upd}}$) & 0.1 & 0.1& 0.1 \\
$\lambda_{\mathrm{da}}$ (Eq.~13, GRL strength) & 0.2 & 0.1  & 0.1  \\
$\lambda_{\text{dist}}$ (Eq.~13, N--$\tilde{N}$ distance) & 0.01 & 0.01 & 0.01\\
$\beta$ (Eq.~14, usage exponent) & 1.0 & 1.0 & 1.0\\
\midrule
Abnormal memory slots $K_{\text{abn}}$ & 100 & 60& 60  \\
Normal memory slots $K_{\text{norm}}$ & 100 & 60 & 60\\
\bottomrule
\end{tabular}
\end{table*}

\section*{S3. Computational Cost}

\subsection*{S3.1 Generation cost}

All pseudo-anomaly and pseudo-normal videos in our experiments are generated with Wan2.2 on a workstation running Ubuntu~24.04.3~LTS equipped with two NVIDIA RTX~6000~Ada~96\,GB GPUs. 
During synthesis, Wan2.2 utilizes approximately 95{,}776\,MiB of GPU memory, and generating a single pseudo-abnormal clip takes about 9~minutes~35~seconds. 
Producing all synthesized clips for ShanghaiTech (140 pseudo-abnormal and 50 pseudo-normal clips) requires roughly 30~hours of wall-clock time, while the full set for UCF-Crime (260 pseudo-abnormal and 80 pseudo-normal clips) takes about 55~hours. 
For XD-Violence, we synthesize 60 pseudo-abnormal clips (10 per each of 6 classes) and 50 pseudo-normal clips, totaling 110 clips; under the same setup, this requires roughly 18~hours of wall-clock time.
Considering the scarcity and privacy constraints of real abnormal footage, this generation cost remains modest; PA-VAD enables efficient large-scale pseudo-anomaly creation, offering a practical and scalable alternative to collecting real anomalies.

\end{document}